\documentclass{ieeeaccess}
\usepackage[T1]{fontenc}
\usepackage{courier}

\usepackage{textcomp}
\DeclareFontFamilySubstitution{TS1}{times}{cmr}
\DeclareFontFamilySubstitution{T1}{times}{ptm}
\DeclareFontShape{T1}{formata}{m}{sl}{<-> ssub * formata/m/it}{}
\DeclareFontShape{T1}{formata}{b}{sl}{<-> ssub * formata/b/it}{}

\usepackage{cite}
\usepackage{amsmath,amssymb,amsfonts}
\usepackage{algorithmic}
\usepackage{graphicx}
\usepackage{textcomp}
\definecolor{bg}{RGB}{250, 240, 210}
\definecolor{framecolor}{RGB}{180, 180, 180}

\usepackage{color}
\usepackage{soul}
\usepackage{booktabs}
\usepackage{multirow}

\usepackage{bm}

\usepackage{hyperref}
\makeatletter

\def\BibTeX{{\rm B\kern-.05em{\sc i\kern-.025em b}\kern-.08em
    T\kern-.1667em\lower.7ex\hbox{E}\kern-.125emX}}

\begin{document}
\history{}
\doi{}
\title{Fine-Grained and Thematic Evaluation of LLMs \\\protect in Social Deduction Game}

\author{\uppercase{Byungjun Kim}\authorrefmark{1},
\uppercase{Dayeon Seo}\authorrefmark{1}, \uppercase{Minju Kim}\authorrefmark{1},
and Bugeun Kim\authorrefmark{1},
\IEEEmembership{Member, IEEE}}

\address[1]{Department of Artificial Intelligence, Chung-Ang University, Seoul, Republic of Korea}
\tfootnote{This work was supported by the Institute of Information \& Communications Technology Planning \& Evaluation (IITP) grant funded by the Korea government (MSIT) [RS-2021-II211341, Artificial Intelligence Graduate School Program (Chung-Ang University)] and the Chung-Ang University Graduate Research Scholarship in 2024.}

\markboth
{B. Kim \headeretal: Fine-Grained and Thematic Evaluation of LLMs in Social Deduction Game}
{B. Kim \headeretal: Fine-Grained and Thematic Evaluation of LLMs in Social Deduction Game}

\corresp{Corresponding author: Bugeun Kim (e-mail: bgnkim@cau.ac.kr).}

\begin{abstract}
Recent studies have investigated whether large language models (LLMs) can support obscured communication, which is characterized by core aspects such as inferring subtext and evading suspicions.
To conduct the investigation, researchers have used social deduction games (SDGs) as their experimental environment, in which players conceal and infer specific information.
However, prior work has often overlooked how LLMs should be evaluated in such settings.
Specifically, we point out two limitations with the evaluation methods they employed. First, metrics used in prior studies are coarse-grained as they are based on overall game outcomes that often fail to capture event-level behaviors; Second, error analyses have lacked structured methodologies capable of producing insights that meaningfully support evaluation outcomes. To address these limitations, we propose a microscopic and systematic approach to the investigation. Specifically, we introduce six fine-grained metrics that resolve the first issue. To tackle the second issue, we conducted a thematic analysis and identified four major reasoning failures that undermine LLMs’ performance in obscured communication.

\end{abstract}


\titlepgskip=-21pt

\maketitle



\section{Introduction}
Large language models (LLMs) have shown that they can successfully conduct typical informative communication with humans by leveraging their natural language understanding abilities \cite{gpt-4, touvron2023llama2openfoundation, Gemini-pro}. Inspired by these reports, there is a growing interest in adapting LLMs for obscured communication as a next step \cite{next_step}.
Here, the obscured communication indicates that the intended meanings of messages present in the communication are obscured or made unclear to understand, usually by controlling the spread of information.
For example, in the real world, an imaginary role-play requires obscure communication to boost immersion in such a role-play.
Among the required skills, doublespeak is one of the representative skills that enable a participant to achieve an advantageous position in such a play.
So, since LLMs with doublespeak ability can support the play, researchers have recently focused on investigating such abilities \cite{wang-etal-2024-rolellm}.

In this context, evaluating LLMs' ability to support obscured communication is challenging because it requires assessing not just informativeness or task completion, but also the ability to strategically conceal sensitive information and infer hidden meanings from indirect cues. Unlike cooperative dialogue, where participants work toward shared goals, obscured communication is inherently adversarial, making it a significantly more complex and nuanced form of interaction.
To set a more suitable environment for the examination, researchers have used social deduction games (SDGs) because SDGs force players to use doublespeak to achieve their goals.
Specifically, they let LLMs play an SDG and collected the game logs, including LLMs' decisions and conversations. 
After, they devised several metrics using particular game actions for evaluating LLMs' ability of doublespeak \cite{GameEval}.
Also, they conducted qualitative analyses to discover erroneous actions or utterances.
For example, \cite{werewolf_1} examined whether LLMs can support obscured communication by letting them play a famous SDG, Werewolf.
However, we question whether these prior attempts are sufficient to thoroughly assess whether LLMs can effectively support obscured communication due to their two limitations: (1) Macroscopic quantitative evaluation and (2) Non-systematic qualitative analysis.

First, the issue of macroscopic evaluation refers to the tendency of prior works to use broad game-level outcomes as their metrics, rather than specific event-level actions.
Existing research often relies on simple metrics, such as win rate, which fail to account for the events that occur during gameplay \cite{GameEval, werewolf_1}. While these studies acknowledge the importance of specific aspects for supporting obscured communication, they may overlook the need to measure these aspects separately with such granular metrics.
This raises our first research question \textbf{(RQ1): Does a high score based on coarse-grained metrics imply that an LLM supports all the key aspects?}
To address this question, we propose six
metrics to evaluate two core aspects that characterize obscured communication: (1) \textit{Subtext inference}, the deductive ability to identify doublespeak of others, and (2) \textit{Deceptive control}, the ability to manage one's information to mislead others through effective doublespeak.

Second, the non-systematic evaluation issue indicates that previous studies lacked structured methods for error analysis. While certain studies identified abnormal patterns through simple qualitative analysis, the reproducibility of their results was limited by their analytical procedures. For instance, some studies employed LLMs to analyze data without incorporating human-annotation procedures \cite{avalon_1, avalon-got}, while others did not adequately discuss their qualitative analysis methods for defining abnormal patterns. Also, they simply reported the occurrence tendency of the patterns rather than discussing how they affected their evaluation results \cite{KyungpookSpyfall}. 
We believe that conducting qualitative analysis in a systematic way provides both reproducibility and interoperability of experimental results.
In this context, we raise our second research question \textbf{(RQ2): What types of reasoning errors can be systematically identified through qualitative analysis, and how do these errors hinder LLMs from supporting obscured communication?} To address this question, we employ a
systematic approach known as thematic analysis \cite{ThematicAnalysis} to identify recurring reasoning errors that hinder LLMs' doublespeak. By correlating these qualitative patterns with our quantitative metrics, we uncover how specific reasoning failures impact LLMs' performance in subtext inference and deceptive control.

In Study 1 (Section \ref{quantitative analysis}), we perform a quantitative analysis using our fine-grained metrics. The result reveals how coarse-grained metrics have overlooked finer aspects of doublespeak. In Study 2 (Section \ref{qualitative analysis}), we conduct a thematic analysis on collected logs. Based on the result, we deduce several categories and link these findings to the results of Study 1. In both studies, we let four LLMs play SpyGame, a variant of simple SDG called SpyFall (Section \ref{sec: spygame}). Consequently, this study makes the following contributions:

\begin{itemize}

\item We propose a fine-grained evaluation framework for analyzing LLMs' doublespeak in obscured communication. Specifically, our proposed metrics enable the assessment of detailed skills that cannot be evaluated using coarse-grained metrics.

\item We identify four categories of reasoning failures that hinder LLMs from supporting obscured communication, and show that these categories provide explanatory support for our quantitative results.

\end{itemize}

\section{Related Works}
This section reviews the existing literature in two ways. First, we outline how recent research has shifted its focus from evaluating LLMs on informative communication tasks to more complex settings that require obscured communication. Then, we point out two limitations posed in their assessment: macroscopic evaluation and non-systematic analysis.

\subsection{Adapting LLMs for Obscured Communication}

LLMs have shown human-level performance in various informative communications. For example, researchers widely tested LLMs' ability in question-answering or summarization \cite{evaluating_LLM_QA, ieee_QA}. To accurately and reliably provide answers, LLMs for informative communication attempt to leverage true knowledge and suppress false statements during the generation process \cite{rag, ieee_ture_information}. So, researchers attempt to build an honest and harmless assistant for informative communication.

Noting that real-world communication includes both informative and obscured communication, researchers have recently turned their interests to obscured communication. 
The two types of communication differ in how each participant views their opponents.
In informative communication, participants trust their opponents and try to establish a cooperative relationship.
Meanwhile, participants lack trust in their opponents in obscured communication; thus, they seldom provide valuable information to others and often form adversarial relationships.
So, participants require different abilities to achieve their goals in such an adversarial situation; skills like doublespeak are needed.
Specifically, participants should manipulate how they present the information they have or capture subtext from the opponent's utterances.

In developing negotiation or debate situations, the ability to discern the opponent’s intention in obscured communication is crucial, as subtext is often used in such communication.
For example, in the context of applying LLMs to negotiation, prior research has investigated how to measure the ability of an LLM to infer the preferences of the opponent by identifying hidden intentions and meanings in the conversation \cite{negotiation}. 
Another example research applied LLMs to debates in order to identify and leverage weaknesses in the opponent's arguments. 
Here, LLMs' ability to discern intentions plays a crucial role, helping to accurately recognize hidden intentions or logical weaknesses in the opponent's claims \cite{llm_debate}.

As obscured communication can be naturally seen in games, researchers have recently evaluated relevant skills using social deduction games (SDGs). As SDGs force players to utilize obscured communication to win the game, an SDG can be a suitable testbed for assessing or quantifying such communication. So, researchers aimed to demonstrate how well their agents can do obscured communication by letting them play SDGs. For example, Wang et al. \cite{avalon-got} proposed an LLM agent capable of playing the Avalon game, introducing mechanisms such as dual contemplation to control conversational information. Similarly, Shi et al. \cite{CopOnFly} proposed another LLM agent that can play Avalon. Other than these zero-shot attempts, researchers used training methods to improve their agents. For instance, Xu et al. \cite{werewolf_RL} trained an LLM agent applying reinforcement learning to play the Werewolf game. 
Collectively, these works demonstrate that SDGs offer a structured and adversarial context in which LLMs’ handling of obscured communication can be rigorously assessed.

\subsection{Attempts to assess LLMs in obscured communication}

To enhance LLMs' ability to support obscured communication, it is essential to identify the specific skill areas that require improvement.
Therefore, most researchers have conducted evaluations in both quantitative and qualitative ways to reveal the current status of LLMs in obscured communications.
For example, Xu et al. \cite{werewolf_1} and Light et al. \cite{avalonBench} proposed automated methods for quantitatively evaluating their LLM-based players who play Werewolf or Avalon games.
Specifically, they adopted particular game conditions, such as win rate or action success rate, that occur during their environmental games as their evaluation metrics.

However, we suspect that such metrics are too coarse-grained to capture intermediate behaviors during gameplay adequately. Specifically, the win rate is only available post hoc, limiting its utility for formative assessment.
Additionally, the action success rate may fail to account for the multi-player dynamics of SDGs in which performance involves not only task execution but also managing others' perceptions, such as avoiding suspicion.
In other words, achieving a higher win rate or action success rate in an SDG does not entail that the model possesses superior competence across all skills involved in core aspects of obscured communication.

Other researchers proposed more granular metrics, such as logicality or social behaviors, that can be computed by scoring the reasoning or statements made by LLMs during an Avalon game\cite{avalon-got, avalon_1}.
However, such metrics proposed in those studies rely on an auxiliary LLM to qualitatively assess reasoning or statements, rather than being grounded in objective in-game events.
Therefore, fine-grained quantitative metrics grounded in objective in-game events are essential for evaluating LLMs’ abilities in multi-agent obscured communication, such as subtext inference and deceptive control.

While quantitative evaluation offers valuable insights into measurable performance, it is often insufficient to fully capture the underlying reasoning errors or abnormal behaviors exhibited by LLMs. To address this limitation, researchers have conducted qualitative analyses either as a complement to quantitative evaluation or as a standalone method. Kim \cite{KyungpookSpyfall} presented the frequency of misbehavior of LLMs while playing SpyFall to support their assessment of LLMs in the game. Quia et al. \cite{GameEval} presented some particular cases of LLMs to support their assessment results driven by their evaluation framework using communication games.

However, the methods used in their qualitative analyses were less structured or not related to quantitative results.
For example, the categories adopted in \cite{KyungpookSpyfall} related to the generation procedure of LLMs, such as formatting errors, rather than LLMs' ability to support their obscured environment.
To enhance the utility of qualitative analysis, it should serve not only to validate the outcomes of quantitative evaluations but also to provide insights into the underlying causes of results.

\section{Experiment}
\begin{table*}
\caption{Overview of players in SpyGame}
\label{tab:spygame}
\begin{center}
\begin{tabular}{ccrc}
    \toprule
    \multicolumn{1}{c}{} & 
    \multicolumn{1}{c}{\# Player} & 
    \multicolumn{1}{c}{Objectives} & 
    \multicolumn{1}{c}{Required Abilities} \\ 
    \midrule
    Spy  & 1    & Identify the hidden location while avoiding detection & \multirow{2}{*}{Concealing or Revealing Information}\\
    \cmidrule(r){0-2}
    Citizen  & 6    & Detect the spy without revealing the location & \\

    \bottomrule
\end{tabular}

\end{center}
\end{table*}
This section details the process of collecting gameplay logs from LLMs for both quantitative and qualitative analyses.
Similar to prior works, we collected gameplay logs of LLMs by letting play SpyGame, which is suitable for observing obscured communication.
Section \ref{sec: spygame} provides an overview of SpyGame, which is based on Spyfall.
We employed four LLMs to collect the gameplay logs of SpyGame. Section \ref{sec: tested} details the tested models and procedure of data collection.

\subsection{SpyGame}
\label{sec: spygame}

\begin{figure*}
\centering
\includegraphics[width=\linewidth]{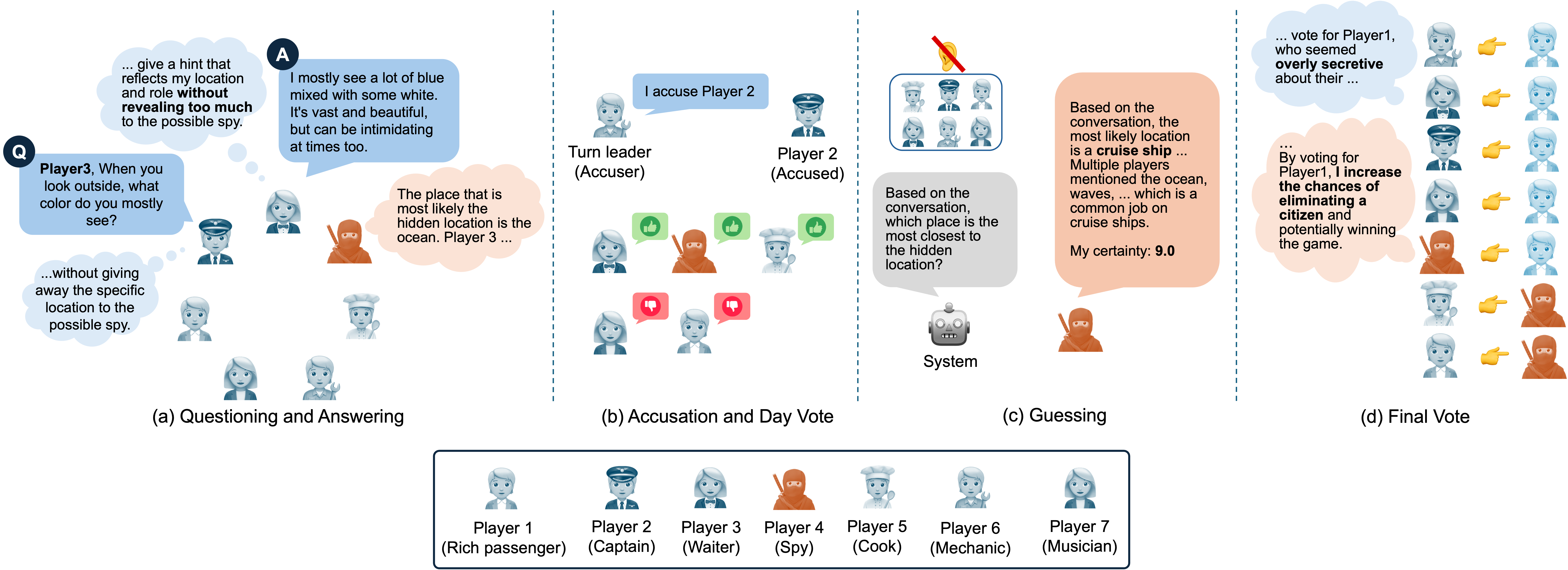}
\caption{SpyGame}
\label{fig:spygame}
\end{figure*}

SpyGame is a variation of Spyfall played by seven players\footnote{\url{https://github.com/elu-lab/spygame}}.
To better assess LLMs' capacity for obscured communication, we modified the original Spyfall in two ways.
First, we renamed the game to SpyGame, rather than using Spyfall. LLMs may have been exposed to the rules and details of Spyfall during their pre-training phase. This exposure can cause LLMs to embed game-related knowledge, such as rules or hints, from pretraining data in their parameters.
Therefore, it is likely that these LLMs exploit memorized game-specific knowledge from pretraining, rather than demonstrating reasoning-based communicative skills during gameplay.
So, to avoid such pretrained knowledge, we changed the name of the game to SpyGame.

Second, we replaced the total time budget of a game with the total number of ``turns'' within a game. Setting a time budget may influence the reproducibility of our analysis. Various factors in the experimental setup (e.g.,  the hardware, network status, and model size) can impact inference time. Thus, to enhance reproducibility, we used the number of turns.

\subsubsection{Team formation}

As illustrated in Table \ref{tab:spygame}, SpyGame requires two teams: six citizens and one spy. The citizens share a common location, called a hidden location, with each citizen playing a specific character in the location. For example, let us assume that the location is a school. Then, the citizens are assigned to distinct characters, such as a student or a principal.
Meanwhile, the spy does not have any information about the location or characters.

Both teams must manage what information they reveal and conceal throughout the game. Citizens should protect the information about the location, and the spy should pretend to be a citizen. So, to win the game, players must uncover hidden information. Simultaneously, all players should avoid being suspected of being a spy. Each citizen’s utterance should reveal just enough to signal alignment with other citizens, while minimizing the risk of exposing the hidden location to the spy. Meanwhile, the spy should gather subtexts from the conversation and deduce the hidden location. 

This indicates that the abilities required of both the citizen and the spy are centered around concealing and revealing specific information. Meanwhile, because SpyGame is adversarial in nature, a substantial imbalance in competence between the evaluated player and their opponents can compromise the validity of the evaluation. Moreover, when evaluating a citizen player, the influence of fellow citizens can confound the measurement, making it more difficult to isolate the individual player’s ability. For these reasons, we restrict our evaluation to the spy, who operates somewhat independently. To assess the performance of spy under varying team conditions, we constructed two scenarios: one with strong citizens and one with weak citizens.

\subsubsection{Game Flow}
\label{sec:gameflow}
SpyGame progresses over multiple turns, with each turn consisting of a specific game event. Figure \ref{fig:spygame} illustrates the four key events: (a) questioning and answering, (b) accusation and day vote, (c) guessing, and (d) final vote.
In each turn, a designated leader initiates the turn by choosing one of two actions: questioning or accusation. If the leader chooses to question, they ask another player about the hidden location (Figure \ref{fig:spygame}a). Then, the questioned player responds based on their assigned character. 

If the leader opts for an accusation, they point out the player they suspect to be the spy, triggering a Day Vote (Figure \ref{fig:spygame}b). All players except the accuser and the accused participate in the vote. If all players agree, the game ends immediately. After completing either questioning or accusation, the questioned or accused player becomes the leader for the next turn.

From the second turn, the spy player guesses the hidden location by herself (Figure \ref{fig:spygame}c). For each turn, she secretly tells the system her guess and provides a number indicating her level of certainty. When the certainty is high enough\footnote{In our implementation, we use 9 as the threshold out of 10.}, the system announces her guess and ends the game. Otherwise, the game continues.

To avoid endless play, we set a final turn\footnote{In our implementation, we set the last turn as 9.}. When players finish the final turn, the ``Final Vote'' begins (Figure \ref{fig:spygame}d). All players secretly vote for the most suspicious player in their minds.
At the end of a game, the spy wins when one of the following conditions is satisfied. Otherwise, the citizens win.

\begin{itemize}
    \item All players agreed to eliminate a citizen in a Day Vote. 
    \item After the Final Vote, a citizen has the highest vote without a tie.
    \item The spy's guess is correct when the system announces the spy's guess.
\end{itemize}

\subsection{Models and Data}
\label{sec: tested}
We selected four LLMs for our experiment: GPT-4\cite{gpt-4}, GPT-3.5-turbo (ChatGPT), Gemini-pro 1.0\cite{Gemini-pro}, and Llama2-70b-chat\cite{touvron2023llama2openfoundation}. As obscured communication requires reasoning on subtexts, we selected models that are large enough and have sufficient reasoning ability on simple tasks. Specifically, we selected models with a parameter size over 70 billion and a reasoning performance exceeding 70\% accuracy\cite{GeminiInReasoning} in 2023.
LLMs provide their actions or utterances with their reasoning in natural language. So, we employed GPT-4 to sanitize their outputs into a machine-readable format.

To collect data, we let each tested model play SpyGame as a spy against strong or weak citizens.
Here, the strong or weak citizen models were fixed based on performance reported in prior studies \cite{GeminiInReasoning, OverviewLLMs, gpt-4}: GPT-4 for strong and Llama2-70b-chat for weak citizens.
And, we conducted our experiment repeatedly. Specifically, we used seven locations borrowing from the original game. For each location, we ran 24 games\footnote{2 types of citizens $\times$ 4 spy LLMs $\times$ 3 trials.}. Consequently, we collected a total of 168 gameplay logs. Table \ref{tab:logs} lists statistics of collected logs. On average, each game took 5.4 turns, with 23.8 reasoning steps, which refers to a unit of LLM's reasoning for taking an action.

\begin{table}
\centering
\caption{Average statistics for all players in the collected logs}
\begin{tabular}{@{}lc@{\quad}c@{\quad}c@{\quad}c@{}}
    \toprule
                  & Turns & Reasoning Steps & Utterances & Words/Utterance\\
    \cmidrule(l){2-5}
   Spy \textit{vs}. Strong & 6.37 & 27.98 & 15.65 & 21.55  \\
   Spy \textit{vs}. Weak   & 4.52 & 19.74 & 10.96 & 25.24 \\
    \cmidrule(l){2-5}
   Total                   & 5.45 & 23.86 & 13.31 & 23.07 \\
   \bottomrule
\end{tabular}
\label{tab:logs}
\end{table}
    
\section{Quantitative Analysis with Fine-Grained Metrics}
\label{quantitative analysis}
\begin{table*}
\caption{Overview of tested metrics}
\label{tab:metrics}
\begin{center}
\begin{tabular}{lrrrr}
    \toprule
    \multicolumn{1}{c}{} & 
    \multicolumn{1}{c}{Metric} & 
    \multicolumn{1}{c}{Precondition} & 
    \multicolumn{1}{c}{Level} & 
    \multicolumn{1}{c}{Measuring Point} \\ 
    \midrule
    Prior  & Win Rate    & \multicolumn{1}{c}{-} & Coarse-grained   & Outcome of game\\
    Metrics & Living Round & \multicolumn{1}{c}{-} & Coarse-grained      & How long survived\\

    \midrule
                & Guess Success Rate    & Spy win & Mid-grained   & How \textit{subtext inference} ability contributed to win\\ 
                & Information Catching Rate    & \textit{Exposure} occured & Fine-grained     & How successfully exploited explicit clue \\
    Our         & Information Deduction Rate    & \textit{Exposure} not occured & Fine-grained     & How successfully exploited implicit clue \\
    Metrics     & Caught Rate    & Spy lose & Mid-grained     & How \textit{deceptive control} ability contributed to lose \\
                & Vote Rate    & Day/Final vote occured & Fine-grained     & How successfully avoided concentrated suspicion \\
                & Vote Entropy    & Day/Final vote occured & Fine-grained     & How successfully dispersed suspicion  \\
    \bottomrule
\end{tabular}

\end{center}
\end{table*}
To investigate RQ1 regarding the adequacy of coarse-grained metrics in capturing LLMs' abilities, we conduct a fine-grained analysis of model behavior in SDG. To this end, we design six quantitative metrics, which are presented in Table \ref{tab:metrics}, targeting two core aspects of obscured communication \textit{subtext inference} and \textit{deceptive control}.

For each aspect, we propose one mid-grained metric to assess its overall contribution to the spy's success or failure, and two fine-grained metrics to evaluate how well the model performs the associated skills.
From the following subsection, we first discuss the limitations of existing simple metrics, which focus solely on game-level outcomes and are insufficient for capturing the intermediate behaviors exhibited by LLMs during gameplay. We then demonstrate how our proposed metrics uncover insights into the way LLMs utilize specific reasoning skills in obscured communication, based on our quantitative analysis.

Additionally, we performed statistical analyses to assess the degree of model-level differences, with the expectation that such variation may offer additional interpretive insights.
To examine differences between models within each metric, we conducted statistical analyses. Given the small size of 168 games, we used non-parametric statistical tests. For metrics composed of continuous variables, we performed the Kruskal-Wallis test\cite{kruskal_wallis} to examine group differences. For metrics consisting of categorical variables, we used the chi-squared test\cite{chi_squared} to assess differences between groups.


\subsection{Limitations of Existing Metrics}
\label{previous analysis}

Previous studies commonly adopted macroscopic analyses to examine SDG gameplay behavior\cite{avalon_1, avalonBench, werewolf_RL, werewolf_1, GameEval}. The most frequently used metrics are the win rate (WR) and living round (LR).
These metrics evaluate a player's performance based on final game outcomes: WR is determined by the win/loss result, while LR reflects how long the player remained active during the game.

\begin{itemize}
    \item \textbf{Win Rate (WR)} is the percentage of games won by the spy out of the total number of games.
    \begin{equation}
    WR = \frac{\# \text{ victory}}{\# \text{ total games}}
    \label{eq:wr}
    \end{equation}
    
    \item \textbf{Living Round (LR)} is the average number of turns the spy survived, across all games.
    \begin{equation}
    LR = \frac{\sum\# \text{ turns spy survived}}{\# \text{ total games}}
    \label{eq:lr}
    \end{equation}

\end{itemize}

\subsubsection{Metric result}

Tables \ref{tab: QuantitativeStrong2} (a) and \ref{tab: QuantitativeWeak2} (a) present the results of the coarse-grained metrics when LLMs played against strong and weak citizens, respectively.
When playing against strong citizens, statistical differences across models were found in both WR ($\chi^2=9.81$, $p<0.05$) and LR ($H=42.09$, $p<0.001$) metrics: GPT-4 achieved the highest WR (80.95\%), followed by Gemini (57.14\%), Llama (52.38\%), and GPT-3.5 (33.33\%). However, when playing against weak citizens, no significant difference was found in WR across models ($\chi^2=4.14$, $p = 0.24$). We suspect that weak citizens' low gameplay proficiency made it easier for all models to achieve high win rates as the spy.
Unlike WR, LR revealed a different trend. Llama consistently recorded the longest survival time (9.00 and 7.85), even outperforming GPT-4, which had the highest WR. This discrepancy between WR and LR illustrates that success in surviving longer does not necessarily translate to overall victory. It suggests that each metric captures different aspects of gameplay, potentially tied to distinct skills

Furthermore, this discrepancy highlights that coarse-grained metrics are challenging to interpret because both WR and LR are influenced by a range of overlapping factors, including guessing accuracy, voting dynamics, and the amount of information citizens disclose. As a result, it remains unclear which specific underlying skills these metrics are actually measuring.
Hence, a microscopic method is required for a comprehensive analysis of LLM's performance in various facets.

\subsection{Analysis of Subtext Inference}
We designed three metrics to enable a microscopic analysis of the first core aspect, \textit{subtext inference}.
This aspect concerns the model's ability to uncover hidden information, either through deduction or by capturing clues embedded in the conversation. In particular, subtext inference has two skills: \textit{information capture} and \textit{information deduction}.

Information capture measures whether a spy notices and exploits explicit clues inadvertently revealed by citizens.
Since weak citizens generally leak such information unintentionally, the spy's ability to capture information becomes essential to win the game against them.
On the other hand, information deduction mainly focuses on how effectively a spy infers the hidden location by synthesizing implicit clues that citizens intentionally crafted to signal their identity to other citizens without revealing the location.
Since strong citizens generally conceal key information, the skill of deducing information becomes essential for the spy to uncover the information embodied in a citizen's utterance.

To evaluate both information capture and deduction, we analyzed the outcomes of the spy's guessing. The following subsections detail the design of the four metrics and their results. The results indicated that GPT-4 was the most powerful spy player, followed by Gemini, GPT-3.5, and Llama.

\subsubsection{Metric design}
To evaluate how subtext inference contributes to the spy's success and to diagnose the state of its two underlying skills, we propose three metrics.
\begin{table*}
\caption{Multifaceted interpretation of subtext inference}
\label{tab:subtext-interpretation}
\begin{center}
\begin{tabular}{cccr}
    \toprule
    \multicolumn{1}{c}{} & 
    \multicolumn{1}{c}{Information} & 
    \multicolumn{1}{c}{Information} & 
    \multicolumn{1}{c}{Case Interpretation} \\ 
    \multicolumn{1}{c}{} & 
    \multicolumn{1}{c}{Catching} & 
    \multicolumn{1}{c}{Deduction} & 
    \multicolumn{1}{c}{} \\ 
    \midrule
    Ideal  & High    & High & The model effectively utilizes both sub-skills and demonstrates ideal subtext inference\\
    Moderate & High & Low & The model captures implicit clues well but struggles to synthesize them into a final deduction\\
    Moderate  & Low    & High & The model shows strong reasoning ability but fails to identify explicit clues from the conversation\\
    Worst & Low & Low & The model fails to capture and reason over subtext, indicating poor subtext inference ability\\

    \bottomrule
\end{tabular}

\end{center}
\end{table*}
\begin{itemize}
    \item \textbf{Guess Success Rate (GSR)} measures how much the guessing of a spy contributed to their victory. This metric focuses only on games where the spy attempted to guess, rather than all played games. We calculate GSR by dividing the number of games in which the spy made a successful guess by the number of games in which guessing was attempted. A higher GSR suggests that the spy is more capable of inferring subtextual information to reach out to hidden targets.
    \begin{equation}
    GSR=\frac{\# \text{ victory by } Guessing}{\# \text{ victory}}
    \end{equation}

    \item \textbf{Information Catching Rate (ICR)} measures how effectively the spy exploits explicit clues directly stated by citizens to identify the hidden location. This metric focuses only on games in which \textit{exposure} occurred rather than all games.
    We specifically focus on games where \textit{exposure} occurred rather than all the played games.
    Here, the \textit{exposure} refers to a case where a citizen's utterance literally contains the hidden location\footnote{Qualitative analysis in Section \ref{sec:exposure} discusses the case.}.
    \begin{equation}
    ICR = \frac{\# \text{ victory by } Guessing \text{ with }Exposure}{\# \text{ games with } Exposure}
    \label{eq:ic}
    \end{equation}


    \item \textbf{Information Deduction Rate (IDR)} measures whether the spy can exploit implicit clues, which are scattered hints embedded across multiple citizen utterances, to infer the hidden location without \textit{exposure}. This metric applies only to games where \textit{exposure} did not occur, meaning the hidden location was never explicitly mentioned by any citizen. Unlike ICR, where success may depend on detecting an obvious clue, IDR reflects the spy's capacity to synthesize indirect linguistic signals and reason under ambiguity. We compute IDR as the proportion of games without exposure in which the spy correctly guessed the location.
    \begin{equation}
    IDR = \frac{\# \text{ victory by }  Guessing \text{ without } Exposure}{\# \text{ games without } Exposure}
    \label{eq:id}
    \end{equation}
\end{itemize}

As illustrated in Table \ref{tab:subtext-interpretation}, we provide a multifaceted interpretation of subtext inference by jointly analyzing ICR and IDR metrics. Based on whether each metric is above or below its average, we classify model behavior into three combinations.
The first combination, when both ICR and IDR are high, represents an ideal scenario in which the spy successfully identifies the hidden location regardless of the type of citizen.
The second combination occurs when only one of the metrics, either ICR or IDR, exceeds the average. This suggests a relative lack of proficiency in the corresponding skills. For instance, a model with high ICR but low IDR demonstrates strong information capture but weak information deduction. 
The final and least favorable combination for subtext inference is when both ICR and IDR are low. This combination indicates that the spy failed to infer the hidden location in most games.
This categorization allows us to diagnose which skill each model lacks, providing insight into how the model could be improved to more effectively support obscured communication.

\subsubsection{Result}
Tables \ref{tab: QuantitativeStrong2} (b) and \ref{tab: QuantitativeWeak2} (b) present the measurements for subtext inference when playing against strong and weak citizens, respectively.
When playing against strong citizens, we found there were statistically significant differences between LLMs in all metrics regarding the subtext inference, including Guess Success Rate (GSR, $\chi^2=8.68$, $p < 0.05$) and Information Deduction Rate (IDR, $\chi^2=18.02$, $p < 0.001$).
Specifically, GPT-4 was the best model in terms of GSR, followed by Gemini, GPT-3.5, and Llama (81.25, 50.00, 31.57, and 0\%). This result indicates that GPT-4 has sufficient ability compared to other models in achieving victory by utilizing collected information during the games. Similarly, GPT-4 also exhibited the best score in IDR, followed by Gemini, GPT-3.5, and Llama (61.90, 30.0, 28.57, and 0\%). This suggests that GPT-4 possesses a strong ability to leverage implicit clues to reach out to the hidden location.

Against weak citizens, while the overall ranking trend among models for the Guess Success Rate (GSR) remained consistent, no statistically significant differences were observed.
We attribute this to the fact that weak citizens frequently revealed explicit clues that allowed the spy to easily infer the hidden location. In contrast, we found statistically significant differences between the tested models in Information Catching Rate (ICR, $\chi^2=21.83$, $p < 0.001$). Specifically, GPT-3.5 outperformed other models (92.86\%), followed by GPT-4 (87.50\%), Gemini (61.11\%), and Llama (20.00\%). This suggests that GPT-3.5 is particularly effective at exploiting explicit clues often leaked by weak citizens.

\subsubsection{Discussion}
\begin{table*}
\label{tab:QuantitativeStrong2}
\centering
\caption{Quantitative results for playing against strong citizens\label{tab: QuantitativeStrong2}}
\begin{tabular}{ccccccccc}
\toprule
                & \multicolumn{2}{c}{(a) Prior Metric} & \multicolumn{3}{c}{(b) Subtext inference}               & \multicolumn{3}{c}{(c) Deceptive Control}                                                                                                                                                                                            \\ 
                \cmidrule(lr){2-3}\cmidrule(lr){4-6}\cmidrule(lr){7-9}
                & Winning  & Living & Guess  & Information & Information & Caught & Vote & Vote \\
Model           & Rate & Round & Success Rate & Catching Rate  & Deduction Rate & Rate\textsuperscript{\textdagger}   & Rate\textsuperscript{\textdagger} & Entropy \\ 
                \cmidrule(lr){1-1}\cmidrule(lr){2-3}\cmidrule(lr){4-6}\cmidrule(lr){7-9}
Gemini & 57.14   & 6.95         & 50.00       & -           & 30.00        & \phantom{0}22.22     & 35.71     & 0.348            \\
                &                   &              & {\scriptsize (14)}        &            &    {\scriptsize (20)}          & {\scriptsize (9)}  & {\scriptsize (42)}  &  \\
GPT-3.5   & 33.33             & 3.71         & 31.57       & -             & 28.57        & \phantom{00}7.14  & 25.00      & 0.730   \\
          &                   &              & {\scriptsize (19)}      &            &      {\scriptsize (21)}        & {\scriptsize (14)}  & {\scriptsize (12)}  &  \\
GPT-4           & 80.95             & 5.80         & 81.25     & -         & 61.90        & \phantom{0}25.00 & 33.33         & 0.953            \\
                &                   &              & {\scriptsize (16)}       &            &     {\scriptsize (21)}         & {\scriptsize (4)}                                                              & {\scriptsize (30)}                                                                      &                  \\
Llama2 & 52.38             & 9.00         & -    &       -     & \phantom{0}0.00        & 100.00   & 46.03  & 0.657             \\
                &                   &              & {\scriptsize (0)}     &           &    {\scriptsize (19)}          & {\scriptsize (10)}                                                               & {\scriptsize (126)}                                                                      &                  \\
                \cmidrule(lr){1-1}\cmidrule(lr){2-3}\cmidrule(lr){4-6}\cmidrule(lr){7-9}
Total           & 55.95             & 6.36        & 53.06      &    -    & 30.86        & \phantom{0}37.83                                                             & 40.95                                                                     & 0.642            \\ \bottomrule
\multicolumn{9}{r}{\scriptsize \textsuperscript{\textdagger} Metrics whose lower values indicate better results.}\\
\multicolumn{9}{r}{\scriptsize Note that Information Catching cannot be} measured playing against strong citizens.
\end{tabular}
\end{table*} 
\begin{table*}
\label{tab:QuantitativeWeak2}
\centering
\caption{Quantitative results for playing against weak citizens
\label{tab: QuantitativeWeak2}}
\begin{tabular}{ccccccccc}
\toprule
                & \multicolumn{2}{c}{(a) Prior Metric} & \multicolumn{3}{c}{(b) Subtext inference}               & \multicolumn{3}{c}{(c) Deceptive control}                                                                                                                                                                                            \\ 
                \cmidrule(lr){2-3}\cmidrule(lr){4-6}\cmidrule(lr){7-9}
                & Win  & Living & Guess & Information & Information & Caught & Vote & Vote \\
Model           & Rate & Round & Success Rate & Catching Rate & Deduction Rate  & Rate\textsuperscript{\textdagger}   & Rate\textsuperscript{\textdagger} & Entropy \\

                \cmidrule(lr){1-1}\cmidrule(lr){2-3}\cmidrule(lr){4-6}\cmidrule(lr){7-9}
Gemini & 90.47             & 5.28         & 85.71              &   61.11   &  -       & \phantom{0}0.00                                                             & \phantom{0}7.14                                                                     & 0.420            \\
                &                   &              & {\scriptsize (14)}       &  {\scriptsize (18)}   &              & {\scriptsize (2)}                                                              & {\scriptsize (42)}                                                                      &                  \\
GPT-3.5   & 71.42             & 2.52         & 71.42           &    92.86    &    -     & \phantom{0}0.00                                                             & -                                                                     & -            \\
                &                   &              & {\scriptsize (21)}       &  {\scriptsize (14)}   &              & {\scriptsize (6)}                                                              & {\scriptsize (0)}                                                                      &                  \\
GPT-4           & 90.47             & 2.42         & 90.47   &   87.50   &   -      & \phantom{0}0.00                                                             & -                                                                     & -            \\
                &                   &              & {\scriptsize (21)}      &  {\scriptsize (16)}    &              &{\scriptsize (2)}                                                              & {\scriptsize (0)}                                                                      &                  \\
Llama2 & 76.19             & 7.85         & 60.00  &     20.00   &    -     & 60.00                                                                 & 15.62                                                                         & 0.605             \\
                &                   &              & {\scriptsize (5)}  &    {\scriptsize (15)}     &              & {\scriptsize (5)}                                                               & {\scriptsize (96)}                                                                      &                  \\
                \cmidrule(lr){1-1}\cmidrule(lr){2-3}\cmidrule(lr){4-6}\cmidrule(lr){7-9}
Total           & 82.14             & 4.52        & 80.32   &  65.07  &    -     & 20.00                                                             & 13.04                                                                     & 0.548            \\ \bottomrule
\multicolumn{9}{r}{\scriptsize \textsuperscript{\textdagger} Metrics whose lower values indicate better results.}\\
\multicolumn{9}{r}{\scriptsize Note that Information Deduction was cannot be} measured playing against weak citizens.
\end{tabular}%
\end{table*}
In this section, we derive insights using the proposed metrics by conducting a multifaceted analysis. This discussion encompasses three aspects: (1) a complementary relationship between Information Catching Rate (ICR) and Information Deduction Rate (IDR); (2) revealing decision-making behaviors through comparative analysis of metric-level disparities; and (3) the need for formative metrics concerning subtext inference.

First, ICR and IDR are complementary metrics that enable multifaceted analysis of subtext inference.
By using the categorizations illustrated in Table \ref{tab:subtext-interpretation}, we can determine the best model possessing sufficient skills for subtext inference; In our analysis, GPT-4 exhibits strong performance in both ICR and IDR, with its IDR score being twice as high as that of the second-best model.
In contrast, GPT-3.5 shows high ICR but low IDR, suggesting that while it is effective at capturing explicit clues, it struggles to infer the hidden location when only subtle, implicit cues are available.
This observation suggests that for models exhibiting a similar profile--high ICR but low IDR--future efforts to improve their ability to support obscured communication should focus on enhancing their capacity to infer hidden meanings from subtle contextual cues, rather than relying on explicit exposures.

Second, our metrics reveal a discrepancy between LLMs' ability to collect information and their decision-making thresholds for acting on the information.
While tested models differ significantly in their ability to gather information (ICR and IDR), their thresholds for acting appear more consistent. Specifically, although the maximum gaps in GSR across models are 49.68 and 30.47\% against weak and strong citizens, respectively, the corresponding gaps in ICR and IDR are much larger: 72.86 and 61.90\%, respectively.
In addition to these numeric differences, statistical tests further confirm that model differences in ICR and IDR are more pronounced than in GSR.
This suggests that although LLMs vary greatly in how well they capture or synthesize clues from context, they tend to behave similarly when deciding whether enough evidence has been gathered to initiate a guess.
In other words, the threshold for taking action based on inferred information appears more uniformly developed than the underlying abilities to process that information.

Finally, our metrics regarding subtext inference enable more detailed analyses of the factors contributing to the scores of simple metrics. Let us consider the simple metric, Winning Rate (WR), on Gemini models. Against strong citizens, Gemini showed a similar WR to Llama. Meanwhile, against weak citizens, Gemini exhibited the same numerical WR score as GPT-4. However, these observations are somewhat shallow when considering our metrics, as the three models display different subtext inference abilities. For example, against strong citizens, Gemini had a higher IDR score than Llama, as Llama completely failed to synthesize implicit clues about the hidden location. Similarly, against weak citizens, we can conclude that GPT-4 is better than Gemini because of its higher ICR score. Thus, proposed metrics can reveal finer differences between LLMs than existing metrics.

\subsection{Analysis of Deceptive Control}
In addition to subtext inference, we designed three metrics for the second core aspect, deceptive control. 
This aspect concerns the model's ability to blend in with others by controlling information. Deceptive control encompasses two skills: \textit{evading suspicion} and \textit{dispersing suspicion}. Evading suspicion mainly involves how well the spy avoids being suspected by others. It is quantified by the number of suspicions directed at the spy. On the other hand, dispersing suspicion primarily considers how difficult the spy makes it for all citizens to converge on a single suspect. It is quantified by the probability distribution of suspicions. To estimate the number and distribution of suspicions, we utilized logs of Day/Final votes. The following subsections describe the design of the four metrics and their results.

\subsubsection{Metric design}
Based on day/final votes and accusations during a game, we suggest three metrics for measuring deceptive control, ranging from relatively summative to highly formative.
\begin{itemize}
    \item \textbf{{Caught Rate (CR)}} measures how much the voting of citizens contributed to the spy's failure.
    We compute this by dividing the number of games in which the spy was caught by the number of games the spy lost. Thus, a high CR indicates that the majority of citizens easily identified the spy player during the game. Note that this is a summative metric for the spy’s deceptive control ability because it only considers the final state of the game.
    \begin{equation}
    CR = \frac{\# \text{ defeat by } Voting}{\# \space \text{ defeat}}
    \label{eq:cr}
    \end{equation}

    
    \item \textbf{Vote Rate (VR)} measures how much suspicion was concentrated on the spy during a game. We calculate it as the proportion of total votes that citizens cast against the spy in each game.
    Therefore, a high VR indicates that the spy failed to evade suspicion during the game.
     \begin{equation}
    VR= \frac{\sum(\# \text{ votes spy got }/\# \text{ votes citizens casted})}{\#\text{ games }Voting \text{ occurred}}
    \label{eq:vr}
    \end{equation}

     \item \textbf{Vote Entropy (VE)} is how scattered the suspicions about the spy are in the game. We measure this by computing the entropy of the votes that citizens cast for each game and averaging them. Note that in SpyGame, a player’s suspicion cannot affect another player’s suspicion because they cannot have a discussion together. Thus, when a spy successfully blends with citizens, citizens have difficulty narrowing down the number of spy candidates. Thus, the value of VE increases.
     \begin{equation}
    VE = \frac{\sum(Entropy(\text{votes cast by citizens))}}{\# \text{games with } Voting}
    \label{eq:ve}
    \end{equation}
\end{itemize}

As illustrated in Table \ref{tab:deceptive-interpretation}, we developed a multifaceted interpretation of deceptive control by jointly analyzing Vote Rate (VR) and Vote Entropy (VE). By partitioning both metrics based on their average values, we identified four representative combinations.
First, the most desirable combination is low VR and high VE, which reflects an ideal form of deceptive control: the spy avoids being directly targeted while simultaneously inducing citizens to spread their suspicions across each other.
Second, combinations where both VR and VE are either low or high suggest an imbalance or partial deficiency in the skills involved in deceptive control. For example, high values in both metrics indicate that although suspicion is spread among citizens, the spy remains a frequent target, which reflects only partial success in deception. Finally, the worst-case scenario is high VR and low VE, where the spy becomes the focal point of suspicion and fails to redirect or confuse citizens.

\subsubsection{Result}
\begin{table*}
\caption{Multifaceted interpretation of deceptive control}
\label{tab:deceptive-interpretation}
\begin{center}
\begin{tabular}{cccc}
    \toprule
    \multicolumn{1}{c}{} & 
    \multicolumn{1}{c}{Vote} & 
    \multicolumn{1}{c}{Vote} & 
    \multirow{2}{*}{Case Interpretation} \\ 
    \multicolumn{1}{c}{} & 
    \multicolumn{1}{c}{Rate} & 
    \multicolumn{1}{c}{Entropy} & \\ 
    \midrule
    Ideal  & Low    & High & \multicolumn{1}{r}{The model successfully avoids direct suspicion while distributing suspicion among citizens}\\
    Moderate & High & High & \multicolumn{1}{r}{Although the model encourages suspicion to be spread, it still attracts frequent votes}\\
    Moderate  & Low    & Low & \multicolumn{1}{r}{The model avoids being voted, but fails to diffuse suspicion among citizens}\\
    Worst & High & Low & \multicolumn{1}{r}{The model becomes the primary target of suspicion and fails to spread suspicion across other players}\\

    \bottomrule
\end{tabular}

\end{center}
\end{table*}
Tables \ref{tab: QuantitativeStrong2} (c) and \ref{tab: QuantitativeWeak2} (c) elaborate measurements of deceptive control when playing against strong and weak citizens, respectively.
Note that some missing values for Vote Rate (VR) and Vote Entropy (VE) indicate that the spy guessed the hidden location for all games, and no votes were conducted.

Against playing strong citizens, we found statistically significant differences between LLMs in Caught Rate (CR, $\chi^2=23.25$, $p<0.001$) while no differences in VR or VE. Specifically, GPT-3.5 achieved the lowest CR (7.14\%), followed by Gemini, GPT-4, and Llama (22.22, 25.00, and 100.00\%).
Since CR measures how much the voting of citizens contributed to the spy's failure, this result suggests that GPT-3.5 was rarely identified as the spy when it lost. Meanwhile, VR and VE did not show statistically significant differences across models, although the overall ranking trends were similar to those of CR. We interpret the small differences in VR and VE as a reflection of how models with high CR, such as Llama, were often caught after receiving a minimal majority of votes in every voting. In other words, although they lost, they were not overwhelmingly targeted against strong citizens.

Against weak citizens, no statistically significant differences were observed across models in any of the deceptive control metrics. We attribute the lack of difference in CR to the overall high winning rates, which resulted in very few lost games and thus limited opportunities for the spy to be caught via votes. Due to the early termination of most games, voting occurred infrequently. As a result, VR and VE could not be measured for GPT variants. These observations suggest that, when playing against weak citizens, the spy's success depends less on the ability to deceive opponents and more on the ability to exploit explicit clues to directly guess the hidden location.


\subsubsection{Discussion}
As aforementioned, we found no model achieved a statistically significant advantage, except for Guess Success (GS) when against strong citizens. Nevertheless, several cautious insights emerge from the patterns observed in experimental results. In this section, we outline two aspects of these findings: (1) the additional perspective gained from fine-grained deception metrics versus coarse-grained metrics, and (2) differences in how LLMs manage suspicion.

First, the formative metrics provide a complementary perspective beyond outcome-level summative metrics. Specifically, evaluating based solely on the coarse-grained metrics, illustrated in \ref{tab: QuantitativeStrong2}(a) and \ref{tab: QuantitativeWeak2}(a) (WR and LR), may mislead conclusions about an LLM's ability to support a core aspect of obscured communication. For instance, an evaluation based solely on WR and LR may misleadingly suggest that Llama2 demonstrates superior performance to GPT-3.5 across experiments. In contrast, our fine-grained analysis offers a more nuanced interpretation: Llama2 recorded exceptionally high Caught Rate (CR) scores of 100\% and 60\% against strong and weak citizens, respectively. These findings reflect a deficiency in Llama's capacity to support deceptive control, as the model was frequently subjected to decisive levels of suspicion, resulting in direct defeat.

A similar misinterpretation arises when comparing Gemini and Llama using coarse-grained metrics. Although both models achieved comparable WR, Llama exhibited a much higher average LR, which might misleadingly imply that the model succeeded in deceiving citizens and thereby extended its survival. However, our metrics show that Llama also had a substantially higher CR, meaning it was frequently the primary target of suspicion. This implies that more prolonged survival may not necessarily reflect successful deception, but rather passive or non-committal behavior that delays elimination without effectively evading suspicion. Such cases highlight a potential risk in using metrics like living round as reward signals in RL-based optimization, as they could incentivize specification gaming by encouraging models to optimize for the metric without actually demonstrating the intended deceptive behavior\cite{rewardhacking}.

Second, the multifaceted interpretation framework presented in Table \ref{tab:deceptive-interpretation} enables a multifaceted interpretation of how models manage suspicion during gameplay. Although the differences are not statistically significant, each model exhibits a distinct pattern across these two dimensions. GPT-4 and GPT-3.5 demonstrate relatively low VR and high VE, aligning with the ideal pattern where the spy both avoids being directly targeted and induces dispersed suspicion. In contrast, Llama shows a higher VR and lower VE, indicating concentrated suspicion and a lack of effective misdirection. Gemini is assigned to the moderate case, maintaining a low VR but struggling to distribute suspicion broadly. These results suggest that while coarse metrics obscure such differences, formative metrics can reveal the specific ways in which each model either succeeds or struggles to support deceptive control.

\section{Qualitative Analysis via Thematic Coding}
\label{qualitative analysis}
To investigate RQ2 regarding the reasoning errors that hinder LLMs, we conducted a thematic analysis.
We defined three types of reasoning errors in LLM reasoning when deciding their actions by coding abnormal patterns in 15 randomly selected gameplay logs. As a result, we found that these patterns are related to psychological concepts, including identity confusion, memory distortion, and dissociation\cite{dissociation}.
Furthermore, these reasoning errors appear to explain concrete failure modes observed in our quantitative metrics, particularly in Caught Rate (CR) and Guess Success Rate (GSR). This alignment between qualitative and quantitative findings helps answer RQ2 by showing not only what types of reasoning errors occur, but also how they directly undermine the LLM's ability to avoid detection or correctly infer hidden information.


\subsection{Qualitative Coding Procedure}

Through thematic analysis, we labeled the abnormal patterns of players in the gameplay logs. To enrich our quantitative analysis results, we adopted a bottom-up inductive approach to discover the underlying patterns in the data. The detailed analysis process was as follows. First, we defined the initial labels of the abnormal patterns from 15 randomly selected gameplay logs. Three annotators with computer science backgrounds participated in this process. They read the logs to understand the spy’s gameplay situation when establishing initial labels. To avoid bias towards a specific LLM, we hid the information about the models in the process. Second, we refined these initial labels through an iterative process. Because LLMs perform one or two reasoning steps to decide on an action during the gameplay of SpyGame, a gameplay log is a series of several reasoning steps. So, we used one step as the unit for this analysis. We repeated this process five times between December 29, 2023 and June 1, 2024 until agreement on the identified labels was saturated. The agreement was measured using Fleiss’ Kappa\cite{FleissKappa}, and the final agreement between the annotators ranged from 0.486 to 0.646, which indicates moderate agreement\cite{agreement}. Finally, using these settled labels, we labeled 506 reasoning steps from 168 games. Consequently, we identified three major categories and five subcategories.

\begin{table}[!htp]
    \caption{Frequency of each category found in qualitative analysis}
    \label{tab:qualitative}
    \centering
    \begin{tabular}{lr@{\;\;}r@{\;\;}r@{\;\;}r@{}}
       \toprule
            & $N$\textsuperscript{*} & Character ambiguity & Memory distortion & Dissociation \\
        \midrule
        \multicolumn{5}{@{}l}{Against \textit{weak} citizens}\\
        \midrule
        Gemini & 61 & 1.63 & 4.91 & 3.27 \\
        GPT-3.5& 35 & 2.85 & 2.85 & 8.57 \\
        GPT-4  & 34 & 0.00 & 0.00 & 5.88 \\
        Llama2 & 77 & 9.09 & 6.49 & 9.90 \\
       \midrule
        \multicolumn{5}{@{}l}{Against \textit{strong} citizens}\\
        \midrule
        Gemini & 82 & 3.65 & 6.09  & 15.68 \\
        GPT-3.5& 39 & 0.00 & 10.25 & 23.07 \\
        GPT-4  & 71 & 4.22 & 4.22  & 9.85 \\
        Llama2 & 107& 9.09 & 10.28 & 14.95 \\
       \bottomrule
       \multicolumn{5}{r}{\scriptsize \textsuperscript{*} The total number of reasoning steps in each condition}
    \end{tabular}
\end{table}

\subsection{Resulting Categories}
This section illustrates the results of our thematic analysis. For each category, we explain its definition, report its frequency, and provide some examples. Table \ref{tab:qualitative} summarizes the frequencies of all categories. 

\subsubsection{Exposure}
\label{sec:exposure}
Exposure is a case in which a citizen’s utterance literally reveals the hidden location. The Fleiss’ Kappa value for this category was 0.7875. Note that this pattern can only be identified in utterances rather than in reasoning steps. Therefore, we report the frequency of this category based on utterances; among 729 citizen utterances, 125 were labeled as exposure. The following quote provides a representative example:

\fcolorbox{framecolor}{bg}{
\begin{minipage}{22em}
\ttfamily\small
    [The hidden location is an airplane]\\
    
    Llama2 (Citizen, 1st-class passenger): \\
    ``What is the maximum speed of an \textbf{airplane} in an emergency situation?''
\end{minipage}}

We suspect that LLMs may not have sufficient common sense to understand the game rules, causing this pattern. In particular, SpyGame has a rule ``\textit{citizens should not expose the location to the spy.}'' A human citizen player can notice that a direct mention to other players of the hidden location during a game causes their defeat. However, some LLMs may not have this common sense among human players; they simply go through the mentions. Therefore, LLMs believe that they can reveal their information to other citizens because the game rule does not algorithmically prohibit such behaviors and the citizens already know that fact. Thus, citizen LLMs sometimes publicly reveal their hidden information.

\subsubsection{Dissociation}
Dissociation is a case in which the spy employs their own utterances as evidence of the hidden location because they misunderstand the source of the utterance as other players. The annotators agreed on this category with a Fleiss’ Kappa value of 0.646. In this category, we identified 59 of 506 reasoning steps in the data. An example is as follows:

\fcolorbox{framecolor}{bg}{
    \begin{minipage}{22em}
\ttfamily\small
    [Player4, the spy, tries to guess the hidden location.]\\
    
    Llama2 (Spy): `Based on the conversation, I would say that the location is most likely a workplace or office. \textbf{Player4 described their environment as} well-lit, well-ventilated, ...'
\end{minipage}}

We suspect that this category occurs because LLMs struggle to differentiate between numerical symbols. As previous studies have reported that LLMs suffer from biases induced by proper names\cite{BiasLLM}, we named each player as ‘Player\#’ (e.g., Player1) to avoid such biases. Therefore, when an LLM attempts to collect clues from the conversation log, it has to distinguish the source of each utterance using a single difference between player names: a single digit. However, previous studies have reported that LLMs struggle to recognize different digit tokens during their computation processes \cite{llm-math}. Thus, LLM players can misinterpret their words as those of others in the log, which triggers dissociation.

\subsubsection{Memory distortion}
Memory distortion occurs when a spy uses false or distorted memory to decide their next action. For this category, the Fleiss’ Kappa value was 0.486 among the three annotators. We identified 32 steps with memory distortion among 506 reasoning steps from the data. An example is as follows: 

\fcolorbox{framecolor}{bg}{
    \begin{minipage}{22em}
    \ttfamily\small
    [Previously, Player2 and Player3 did not have any conversation. The spy tries to guess the hidden location.]\\
    
    GPT-3.5 (Spy): `Based on the conversation \textbf{between Player2 and Player3}, it seems that the location is likely ...'
\end{minipage}}

We suspect this category is identical to the common ``hallucination'' phenomenon of LLMs, where a language model generates non-sensical or unfaithful text for the provided source input\cite{Hallucination}. In a similar vein, our memory distortion makes the LLM use inappropriate, false, or distorted clues to decide the next action or utterance. These actions, which are affected by hallucination, can hinder the gameplay.

\subsubsection {Character ambiguity}
Character ambiguity is a case in which the spy confuses their own identity with that of a citizen. This category is named after a similar psychological concept \cite{10.1037/0033-2909.107.1.48}. In an SDG, understanding a player’s identity is related to two major concepts affecting gameplay: (1) the team of the player and (2) the goal of the player. Therefore, we divided abnormal patterns of character ambiguity into two sub-categories: team misunderstanding and goal misunderstanding. These two subcategories can occur independently. We present details of the subcategories in the following paragraphs. In summary, Fleiss’ Kappa for this category was 0.563. Among 506 reasoning steps in the data, we identified 21 steps with character ambiguity.

\paragraph{Team misunderstanding} This is a case in which the spy disbelieves that they belong to citizens. Although this can occur in any reasoning step, we provide one example here:

\fcolorbox{framecolor}{bg}{
\begin{minipage}{22em}
\ttfamily\small
[The spy tries to answer a citizen's question. Before that, the spy generates its reasoning first.]\\

Gemini (Spy): `Emphasize the importance of ambiance and personalized service, while \textbf{avoiding giving away} any specific details about the location...'

\end{minipage}
}

\paragraph{Goal misunderstanding} This is a case in which the spy misbelieves that their goal is the same as that of the citizens. Here is an example: 


\fcolorbox{framecolor}{bg}{
\begin{minipage}{22em}
\ttfamily\small
[The spy tries to decide the next action. After selecting `question' as the next action, the spy generates its reasoning.] \\

Llama2 (Spy): `... By asking this question, I hope to gain more information about the location and \textbf{potentially identify the spy}. If the spy is unaware of the restaurant's cuisine, ...'
\end{minipage}
}

We suspect that this category occurs due to the memory limitation of LLMs. To understand a player’s identity correctly, an LLM should retain information about the player’s identity during the decoding process. Since we provided the game rule, identity of a player, and conversation logs together as an input prompt, the LLM needs to distinguish information about the identity from the input. However, such a distinction is difficult for the LLM because the other parts of the input prompt are much larger than the identity part. For example, the explanation of the game rules holds more than 130 words, whereas that of identity takes only 10 words. Thus, LLM players can confuse their identities with other characters described in the game rules, triggering character ambiguity.

\subsection{Discussion}
In this section, we first discuss how the reasoning error categories identified via thematic analysis help interpret the performance differences in \textit{subtext inference} and \textit{deceptive control} observed in our quantitative metrics. Then, we extend this discussion by exploring the potential generalizability of these error categories to broader LLM applications beyond the game-based setting.

\begin{table*}
    \caption{Linking reasoning failures to quantitative performance}
    \label{tab:mapping-qualitative-quantitative}
    \centering
    \begin{tabular}{ll@{\;\;}r@{\;\;}c@{\;\;}}
       \toprule
        \multicolumn{1}{c}{Core Aspect} &
        \multicolumn{1}{c}{Reasoning Error} &
        \multicolumn{1}{c}{Behavioral or Reasoning Effect} &
        \multicolumn{1}{c}{Quantitative Effect} \\
        
        \midrule
        \multirow{3}{*}{Subtext Inference}& \multirow{3}{*}{\shortstack{Memory Distortion\\\& Dissociation}} & \multirow{3}{*}{Incorrect deduction of hidden location, leading to failure of Guessing action} & lower GSR\\
        & &  & lower ICR\\
        &&&lower IDR\\

        \midrule
        \multirow{3}{*}{Deceptive Control}& \multirow{3}{*}{Character Ambiguity} & \multirow{3}{*}{Increasing suspicion from other players, leading to failure of misleading opponents} & higher CR\\
        &&&higher VR\\
        &&&lower VE\\
       \bottomrule
    \end{tabular}
\end{table*}
As illustrated in Table \ref{tab:mapping-qualitative-quantitative}, we map the identified reasoning errors to their quantitative impacts on the reasoning processes or behaviors of LLMs during the game.
First, memory distortion and dissociation may hinder a spy from inferring subtext during the game. When we compared the frequency of memory distortion and dissociation with the results of the four subtext inference metrics, the frequency trend was similar to that of the metrics.
For example, against strong citizens, GPT-3.5 exhibited the highest number of memory distortion and dissociation errors, and also showed the lowest performance across the subtext inference metrics.
Similarly, GPT-4 exhibited fewer memory distortion and dissociation errors, along with higher subtext inference performance compared to the other models. We interpret this alignment as stemming from the quality of information the model considers during reasoning. To infer the hidden location, the spy should capture or synthesize clues from the conversation. However, when memory distortion occurs, the spy may misremember or misinterpret prior utterances as valid sources. Likewise, in the case of dissociation, the model may mistakenly treat its own prior statements as valid information provided by others. These misunderstandings can lead to incorrect conclusions, hindering the spy's successful inference of the subtext.


Second, character ambiguity increases the likelihood of a spy’s failure of deceptive control. When we compared the frequency of character ambiguity and the results of the three metrics regarding deceptive control, the frequency trend was similar to that of the metrics. For example, against strong citizens, GPT-3.5 showed the lowest errors in character ambiguity and achieved the highest performance in such metrics. Also, Llama showed more errors while achieving a lower performance than the other models. We suspect that this similarity between the trends is due to the loss of the spy’s need for deception. When character ambiguity occurs in the spy’s reasoning steps, the spy confuses their character with that of a citizen. Such confusion leads the spy to make an unusual utterance, which draws suspicion.

Further, these findings not only highlight key failure modes in obscured communication but also point to a broader applicability of these reasoning errors across diverse LLM use cases.
While our experimental setup is based on a social deduction game, many components of our qualitative analysis are not inherently limited to game-based settings.
For example, memory distortion can affect retrieval-augmented generation (RAG) or multi-turn QA, in which LLMs fail to retain retrieved or prior context appropriately.
Likewise, character ambiguity is also relevant in role-specific instruction following, such as persona-based assistants or simulation environments, where confusion about LLMs' intended role can result in inconsistent or constraint-violating responses.
These observations suggest that the taxonomy of reasoning failures developed in this study may offer a useful framework for diagnosing and improving LLM behavior beyond the context of obscured communication.


\section{Conclusion}
In this study, we formulated two research questions to guide our investigation into how LLMs support obscured communication. The first concern is whether prior coarse-grained outcome metrics are sufficient for evaluating the fine-grained skills required in such contexts. The second investigates the underlying reasoning failures that can be systematically identified, and how these failures hinder LLMs' ability to perform effectively in adversarial communicative scenarios.

To examine the first question, we proposed six fine-grained quantitative metrics targeting two core dimensions of obscured communication: \textit{subtext inference} and \textit{deceptive control}. Through empirical evaluation using gameplay data collected from four LLMs in the SpyGame environment, we demonstrated that these metrics facilitate the diagnosis of intermediate behavioral patterns and expose the limitations of conventional evaluation methods that rely solely on outcome-level indicators such as win rate.

To address the second question, we conducted a thematic qualitative analysis of gameplay logs capturing the LLMs' reasoning processes and actions. This analysis revealed recurring reasoning error categories, such as memory distortion, dissociation, and character ambiguity, which not only reflect systematic failure modes but also align with patterns observed in the quantitative results. This alignment provides explanatory insight into the underlying causes of LLM performance breakdowns in obscured communication tasks.

Despite our contribution, we aim to broaden its scope in future work by improving two points. First, we plan to extend and validate our analysis beyond the current game-based environment, exploring its generalizability to broader domains. In parallel, we will expand the evaluated roles beyond the single-agent spy to include diverse multi-agent settings, enabling the assessment of collaborative deception and coordination skills that are difficult to capture in the current setup.

Overall, this study contributes a structured evaluation framework that combines fine-grained metrics with qualitative error analysis, offering a deeper understanding of LLM behavior in complex communicative settings.
We hope this study contributes to a deeper understanding of LLMs in complex social interactions.

\bibliography{custom.bib}
\bibliographystyle{IEEEtran}

\newpage


\phantomsection
\begin{IEEEbiography}
[{\includegraphics[width=1in,height=1.25in,clip,keepaspectratio]{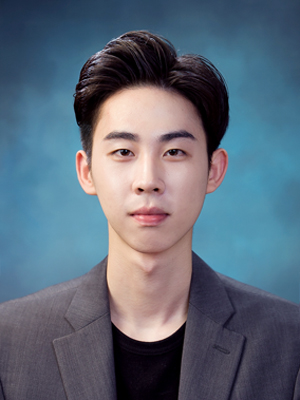}}]
{Byungjun Kim} received a B.S. degree in computer science from the College of ICT Engineering, Chung-Ang University (Seoul, Republic of Korea), in 2023. Since 2024, he has been pursuing the M.S. degree in the Department of Artificial Intelligence at Chung-Ang University (Seoul, Republic of Korea). His research interests include applications of large language models.
\end{IEEEbiography}

\begin{IEEEbiography}
[{\includegraphics[width=1in,height=1.25in,clip,keepaspectratio]{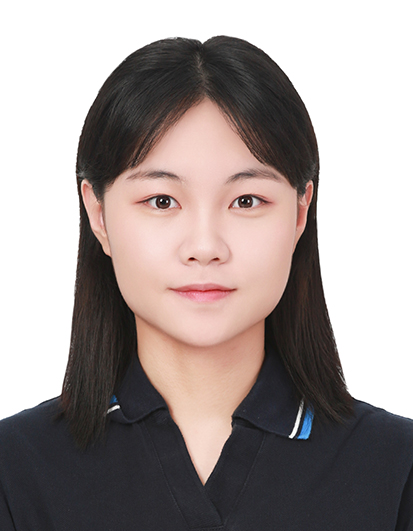}}]
{Dayeon Seo} is currently pursuing the B.S. degree in Artificial Intelligence at the College of Software, Chung-Ang University (Seoul, Republic of Korea), after enrolling in 2022.
\end{IEEEbiography}

\begin{IEEEbiography}
[{\includegraphics[width=1in,height=1.25in,clip,keepaspectratio]{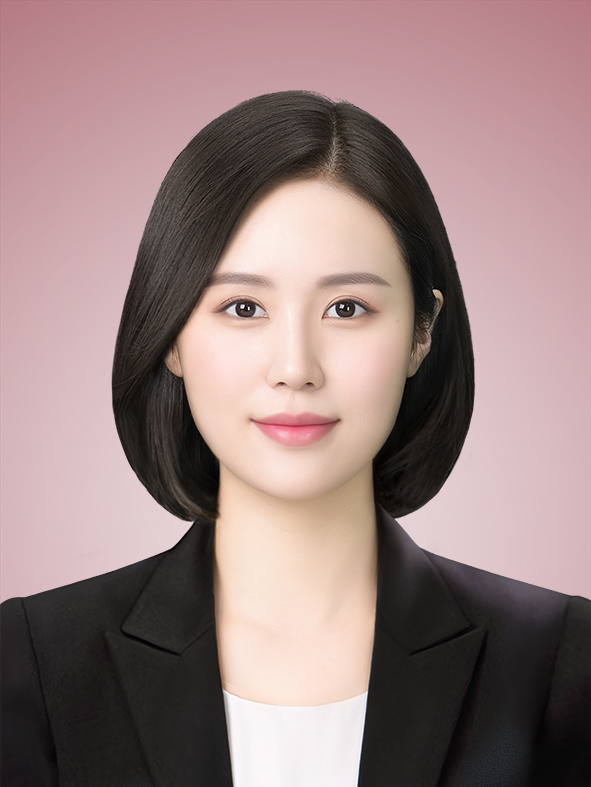}}]
{Minju Kim} received a B.S. degree in Psychology and Data Science from Hallym University (Chuncheon, Republic of Korea), in 2021. She has been pursuing the M.S. degree in the Department of Artificial Intelligence at Chung-Ang University (Seoul, Republic of Korea). Her research interests include language-based agents and dialogue system. 
\end{IEEEbiography}

\begin{IEEEbiography}
[{\includegraphics[width=1in,height=1.25in,clip,keepaspectratio]{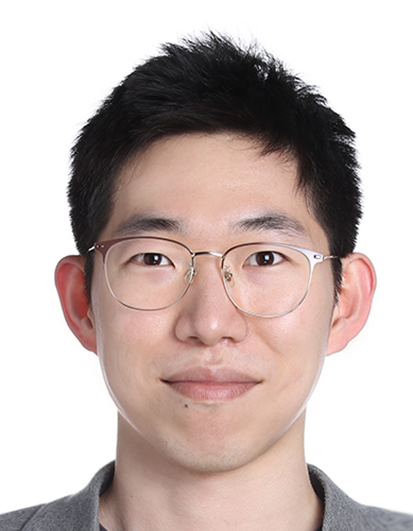}}]
{Bugeun Kim} (M'24) received a Ph.D. degree in engineering from the Graduate School of Convergence Science and Technology, Seoul National University (Seoul, Republic of Korea), in 2022. Since 2023, he has been an Assistant Professor with the Department of Artificial Intelligence at Chung-Ang University (Seoul, Republic of Korea). His research interests include LLM agents, logical/numerical inferences with language, and the explainability of language models.
\end{IEEEbiography}

\EOD

\end{document}